\pdfoutput=1

\documentclass[11pt]{article}

\usepackage{EMNLP2023}
\usepackage{tabularx}
\usepackage{times}
\usepackage{latexsym,amsmath}
\usepackage{amsfonts}  

\usepackage{bm, bbm}
\usepackage{algorithm, algorithmic}
\newtheorem{theorem}{Theorem}
\usepackage[T1]{fontenc}
\usepackage{graphicx}

\usepackage[utf8]{inputenc}
\usepackage{multicol}

\usepackage{microtype}
\usepackage{caption}

\usepackage{inconsolata}

\usepackage{booktabs}

\title{Effective Proxy for Human Labeling: Ensemble Disagreement Scores in Large Language Models for Industrial NLP
}

\author{Wei Du, Laksh Advani, Yashmeet Gambhir,\\ \textbf{Daniel Perry}, \textbf{Prashant Shiralkar}, \textbf{Zhengzheng Xing}, \textbf{Aaron Colak}\\
Qualtrics, Seattle\\
\{weidu, ladvani, yashmeetg, dperry, pshiralkar, zxing, aaronrc\}@qualtrics.com}

\begin{document}
\maketitle
\begin{abstract}
Large language models (LLMs) have demonstrated significant capability to generalize across a large number of NLP tasks.  For industry applications, it is imperative to assess the performance of the LLM on unlabeled production data from time to time to validate for a real-world setting.  Human labeling to assess model error requires considerable expense and time delay.  Here we demonstrate that ensemble disagreement scores work well as a proxy for human labeling for language models in zero-shot, few-shot, and fine-tuned settings, per our evaluation on keyphrase extraction (KPE) task. We measure fidelity of the results by comparing to true error measured from human labeled ground truth.  We contrast with the alternative of using another LLM as a source of machine labels, or ‘silver labels’. Results across various languages and domains show disagreement scores with a mean average error (MAE) as low as 0.4\% and on average 13.8\% better than using silver labels to measure performance.

\end{abstract}

\section{Introduction}
We have recently seen significant progress on many natural language processing (NLP) tasks using the latest generative pretrained models such as GPT \cite{openai2023gpt, ouyang2022training}, PaLM \cite{chowdhery2022palm}, and many others \cite{touvron2023llama, bai2022constitutional, penedo2023refinedweb, alpaca}.  This new generation of models opens up many new possibilities including competitive performance in zero-shot and few-shot settings for tasks that have typically been modeled using a supervised setting \cite{openai2023gpt}.  More established language models (BERT \cite{devlin2019bert}, RoBERTa \cite{liu2019roberta}, XLM-Roberta \cite{conneau2020unsupervised}, etc.) provide a strong balance of inference cost and task performance for such systems.  This broad class of large language models (LLMs) used for complex supervised NLP tasks share the problem of how to effectively assess performance in production settings where we don’t yet have human labels due to cost or urgency.  

The ability to judge model capability becomes important for production settings where we often have to decide whether to launch a model in a new domain or for a new language where we have few or no labels ready.  This is also known as few-shot and zero-shot performance, respectively.  Scaling models up to new domains and new languages quickly becomes an expensive proposition in terms of labeling.  For example, if we have two new domains and ten languages, this results in twenty new label sets that need to be generated.  Having the capability to guide that investment or possibly eliminate the need for extensive human labeling for some subset of those domains/languages becomes very valuable. 

There have been many approaches to assess the performance of LLMs without human labels, including efforts to assess the performance of task-specific models.  \cite{kamath-etal-2020-selective} explored evaluating fine-tuned question answering models on out of domain data, relevant to question answering problems. More recently, \cite{fu2023estimating} creates a meta-model responsible for predicting the accuracy of the LLM model using the model’s confidence scores as features.  Methods from the computer vision (CV) domain to assess unlabeled data more generally have, for example, proposed the average threshold confidence method that learns a threshold over the model’s confidence, predicting accuracy as the fraction of unlabeled examples exceeding that threshold \cite{DBLP:conf/iclr/GargBLNS22}, or iteratively learn an ensemble of models to identify misclassified data points and perform self-training to improve the ensemble with the identified points \cite{chen2021detecting}. However, the metrics and hyperparameters in previous works are specifically for classification tasks and cannot be easily extended to more complex tasks. 

We propose adapting \emph{disagreement scores} in \cite{JiangNBK22, kirsch2022note}, also from the CV domain, to assess model quality for these supervised NLP tasks.  A \emph{disagreement score} is computed by first training a \emph{well-calibrated} ensemble of models and then measuring how similar their respective predictions are on the same input. The intuition is that models will agree on highly confident (likely correct) predictions and disagree on less confident (likely wrong) predictions.
One way to develop a \emph{well calibrated} ensemble is to train the same model on the same dataset but changing initial random seed among the ensemble members, as proposed in \cite{JiangNBK22} for the CV domain.  

In this paper, we adapt the same approach for the NLP tasks to understand the prediction performance across different domains (survey responses, conversation text, and social media chats) and languages. Inspired by the latest work on LLMs, as another alternative to human labeling, we explore leveraging a few-shot GPT-4 as an oracle model to provide a ‘silver label’.  We find that disagreement scores of a well-calibrated ensemble work better at predicting a single model's performance for a complex keyphrase extraction task (KPE) than GPT-4 as an oracle model. Our evaluation comparing XLM-Roberta \cite{conneau-etal-2020-unsupervised}, GPT-3 \cite{brown2020language}, and GPT-4 models \cite{openai2023gpt} shows that disagreement scores provide  estimation of model performance with mean average error (MAE) as low as 0.4\% and on average 13.8\% better than using silver labels. 


\section{Approach: Assessing error without human labels}

\subsection{Adapting Disagreement for Natural Language Tasks}
\label{sec:adapt}

We define $\mathcal{D}$ be a distribution over $\mathcal{X} \times \mathcal{Y}$, where $\mathcal{X}$ is the space of input features to the model and $\mathcal{Y}$ the space of output values from the model. Let $(X, Y)$ denote the random variable from $\mathcal{D}$ and $(x, y)$ be sampled values taken from $\mathcal{D}$. Let $h: \mathcal{X} \rightarrow \mathcal{Y}$ denote a hypothesis from a hypothesis space $\mathcal{H}$. We assume $\mathcal{A}$ be a stochastic training algorithm that induces a distribution $\mathcal{H}_{\mathcal{A}}$ from $\mathcal{H}$. Let $h \in \mathcal{H}_{\mathcal{A}}$ and $h' \in \mathcal{H}_{\mathcal{A}}$ be two random hypotheses output by two independent runs of the training algorithm $\mathcal{A}$. 
We denote the test error and disagreement score for $h \in \mathcal{H}_{\mathcal{A}}$ and $h' \in \mathcal{H}_{\mathcal{A}}$ over $\mathcal{D}$ as the following:
\begin{equation}\label{eq:testerror}
    Test^{err}_{\mathcal{D}}(h) = \mathbb{E}_{\mathcal{D}}[h(X) \neq Y]
\end{equation}
\begin{equation}\label{eq:disagreement}
    Dis_{\mathcal{D}}(h, h') = \mathbb{E}_{\mathcal{D}}[h(X) \neq h'(X)]
\end{equation}
The relationship between $Test^{err}_{\mathcal{D}}(h)$ and $Dis_{\mathcal{D}}(h, h')$ is described as the following Theorem \ref{thm: GeneralizationDisagreement} \cite{JiangNBK22}. 
\begin{theorem}\label{thm: GeneralizationDisagreement}
Given a stochastic learning algorithm $\mathcal{A}$,if its corresponding ensemble satisfies class-wise calibration, then we have:
\begin{equation}
   \mathbb{E}_{h, h' \sim \mathcal{H}_{\mathcal{A}}}[Dis_{\mathcal{D}}(h, h')]
   =   \mathbb{E}_{h \sim \mathcal{H}_{\mathcal{A}}}[(Test^{err}_{\mathcal{D}}(h)).
\end{equation}
\end{theorem}
In this paper, we focus on a sequence-to-sequence task, keyphrase extraction (KPE). 
We use the F1 score instead of test error to measure model quality and agreement instead of disagreement to measure model disparity.  These choices are justified due to the mathematical relationship of model error to F1 score and agreement to disagreement (see Appendix~\ref{sec:f1_error}).  For the computation of KPE agreement, for each sentence we extract the keyphrases using the two models and compute the agreement score as the ratio of common keyphrases extracted to the total number of keyphrases extracted.  The disagreement score is simply $1 - \alpha$, where $\alpha$ is the agreement score.

To estimate model error on unlabeled data, we first train a set of KPE models using different random seeds on the training set.  Then we compute both the disagreement score and the error on a labeled test set to collect all data pairs (F1 score, agreement score). Based on these data pairs, we fit
a simple linear regression model for error prediction, similar to that employed in \cite{JiangNBK22}.



\subsection{LLM as an Oracle}

We have witnessed impressive performance of recent LLMs like GPT-4 on a wide variety of tasks in a zero-shot manner, leading to an increased demand and interest in using them as both a label source for testing data as well for their representation abilities.  Utilizing a model for labeling can result in significant costs savings \cite{tornberg2023chatgpt}.  We include labeling from few-shot prompted GPT-4 as an alternative approach to measure model performance.

\section{Models and Data}

\subsection{Models and Tasks}
We explore using three types of models, all trained for the same KPE task: XLM-Roberta , GPT-3, and GPT-4. The KPE task is representative of many typical industrial NLP tasks, because it is a fundamental and complex problem \cite{song-etal-2023-survey}.  The KPE task consists in taking an input text and and producing a set of textual spans, if any, representing keyphrases as output, which is typically modeled as a sequence to sequence model. 
Consistent with existing approaches \cite{JiangNBK22}, we use mean absolute error (MAE), as the primary metric for measuring fidelity of a proxy error method to the true error measured against human label ground truth.  In this case 
\begin{equation}
\text{MAE} = \frac{1}{n} \sum_{i=1}^{n} | \text{err}^{\text{proxy}}_i - \text{err}^{\text{true}}_i |,
\end{equation} where $\text{err}^{\text{proxy}}_i$ is the proxy or approximated error of the model for the $i$-th experiment and $\text{err}^{\text{true}}_i$ the corresponding true error based on ground truth data.



\subsection{Datasets}\label{sec:dataset}  
We evaluated our approach on three internal datasets corresponding to three distinct domains namely, survey response data, Twitter data, and recorded customer service conversations. The survey-response data is a corpus of 98,844 pairs of survey questions with their appropriate textual responses across 10 languages which we refer to in standard language abbreviations, see Table~\ref{tab:survey-data-statistics} in the appendix for details.  We reserve 79,634 pairs as training and validation data and the other 19,210 as testing data. The Twitter data corpus and the customer support corpus are a collection of 500 tweets relating to customer support and 500 customer service conversation threads respectively.





\section{Experimental Results and Analysis}

We evaluated the the disagreement scoring approach for the KPE task on 10 different languages and three domains using the three models: XLM-R, a fine-tuned GPT-3, and a few-shot prompted GPT-4 model. In the following two sections, we look at evaluations when languages and domains are held out during fine-tuning. In \ref{sec:gpt-4-silver-label}, we look at the case when GPT-4 is used as an oracle for ground truth in a zero-shot manner, without any fine tuning. Table \ref{tab:Language-change} shows a summary of the results on the anonymized survey data.

\begin{table}[htbp]
\centering
\scalebox{0.6}{
\begin{tabular}{llllll}
\hline
\textbf{Language} & \textbf{Avg F1} & \textbf{Avg Predicted F1} & \textbf{MAE} \\
\hline
XLM-R-JA & 0.567 & 0.530 & 0.037 & \\
XLM-R-FR & 0.765 & 0.781 & 0.016 \\
XLM-R-KO & 0.714 & 0.721 & 0.007 \\
Curie-JA-ALL & 0.160 & 0.448 & 0.288 \\
Curie-FRA-ALL & 0.674 & 0.577 & 0.097 \\
Curie-KO-ALL  & 0.395 & 0.305 & 0.080 \\
Curie-FR-EU  & 0.674 & 0.639 & 0.035 \\
Curie-ES-EU & 0.441 & 0.443 & 0.002 \\
GPT-4-EN & 0.427 & 0.595 & 0.168  \\
GPT-4-ES & 0.319 & 0.301 & 0.018 \\
GPT-4-FR & 0.596 & 0.426 & 0.170 \\
GPT-4-IT & 0.356 & 0.373 & 0.017 \\
\hline
\end{tabular}}
\caption{Prediction performance of language change for XLM-R, Curie and GPT-4. Avg F1: average groundtruth F1; Avg Predicted F1: average predicted F1 from fitted linear function.}
\label{tab:Language-change}
\end{table}

\subsection{Language change for LLM}\label{sec:lang-change}


\noindent\textbf{XLM-R}. We fine-tuned the XLM-R base models, with 125M parameters, on all 10 languages with anonymized survey data (Section \ref{sec:dataset}). For each language, we trained four models on that language using the same data but with different seeds, recording F1 scores on the respective language-specific test data. We compute the disagreement score with the other models, giving us six total disagreement scores per language which are then averaged to arrive at the average disagreement score per language. Since we have 10 languages and 4 models, we have 40 (F1 score, disagreement score) pairs for making a prediction. Taking JA as an example, we we use the other 9 languages (36 points) to fit the curve and derive its final prediction (F1 score) as $y = 0.809 x + 0.09631$, where $x$ is the agreement score variable. The MAE for JA is then 3.7\% (first row in Table \ref{tab:Language-change} denoted as XLM-R-JA). 

\noindent\textbf{Curie}.
We use the same training data as XLM-R to fine-tune a GPT-3 model with 13B parameters, known as Curie, through the API provided by OpenAI.\footnote{https://platform.openai.com/docs/guides/fine-tuning}. To understand Curie's performance on Asian vs. all languages, we consider two scenarios: one only focusing on European (EU) languages, and second with all the languages (EU + Asian languages). 


\noindent \textbf{GPT-4}.
We explored using zero-shot and various sizes of few-shot training for GPT-4 and found that 100-shot training did the best. We randomly sample 100 data records from the anonymized survey data for each language for prompting, and use the same test data as used for XLM-R and Curie. The results in Table \ref{tab:Language-change} are using 100-shot prompting and our experiments were limited to EN, ES, FR, and IT due to time constraints.

We make the following observations. First, all LLMs, whether fine-tuned or used as zero-shot, are bounded by 12.9\% MAE on average, encouraging their use for labeling and evaluation needs. The average performance of XLM-R is 2.49\% MAE using all 10 languages  (XLM-R-All), 2.39\% using EU-only (XLM-R-EU), that of Curie is 12.9\% MAE using all languages (Curie-All), 2.09\% using EU-only (Curie-EU), while GPT-4 has 9.38\% MAE using the 4 languages tested. Second, comparing performance on subsets of languages, we find that LLMs struggle on Asian languages, likely due to the differences in pre-training corpora and our test datasets. Finally, LLMs like GPT-4, when used in zero-shot manner, lead to suboptimal performance as compared to ones that are fine-tuned.



\subsection{Domain change for LLM}
We used a test set based on Twitter data and anonymized conversation (conv) data for testing disagreement scoring approach across different domains. We had both datasets annotated by our internal professional annotators and compared the predicted F1 scores from the XLM-R, Curie and GPT-4 models with the actual F1 scores from the human annotations. Table \ref{tab:domain-change} shows the results.

\begin{table}[htbp]
\centering
\scalebox{0.75}{
\begin{tabular}{llllll}
\hline
\textbf{Language} & \textbf{Avg F1} & \textbf{Avg Predicted F1} & \textbf{MAE}\\
\hline
XLM-R-conv & 0.647 & 0.669 & 0.022 \\
XLM-R-Twitter & 0.370 & 0.452 & 0.082 \\
Curie-conv-EU & 0.286 & 0.255 & 0.031 \\
Curie-Twitter-EU  & 0.210 & 0.271 & 0.061 \\
GPT-4-conv & 0.368 & 0.476 & 0.108 \\
GPT-4-Twitter & 0.292 & 0.459 & 0.167 \\
\hline
\end{tabular}}
\caption{Prediction performance of domain change for XLM-R, Curie and GPT-4. Avg F1: average groundtruth F1; Avg Predicted F1: average predicted F1 from fitted linear function. }
\label{tab:domain-change}
\end{table}
\vspace{-0.6cm}
First, the prediction performance of XLM-R and Curie models on conv and Twitter data is better as compared to GPT-4 models, with an average of 4.9\% MAE vs. GPT-4's average of 13.8\% MAE. It is not surprising because XLM-R and Curie have more data points to fit the prediction function, making them more accurate. Note that we only used data points from European languages for Curie due to the distribution gap we observed in Asian languages in Section \ref{sec:lang-change}. Second, the average MAE of the conv data across all three models is 5.3\%, which is lower than that for Twitter data having 10.3\% MAE. We conjecture that this is likely due to the fact that Twitter data is much more noisy, indicating larger domain shift. 



\subsection{GPT-4 few-shot prompt silver label for XLM-R and Curie}\label{sec:gpt-4-silver-label}
To study how well GPT-4 can be used as a silver label generator for the KPE tasks, we fine-tuned a XLM-R model and a Curie model.  We measured error using human labels referred to as \emph{gold labels} and measured error using GPT-4 generated labels or \emph{silver labels}, summarized in Table~\ref{tab:silver-labelling}. Appendix \ref{sec:gpt-4-prompt} shows how we prompt GPT-4 models.

Overall, we observe poor prediction capabilities using 100-shot GPT-4 as a label source.  With XLM-R, we observe a MAE of 31.3\%, 29.1\%, 10.4\%, and 19.3\% for EN, ES, FR and IT respectively. For a practitioner, this MAE is too high to make a confident decision about whether a language requires more human training labels or whether a model is ready for launch. 
For Curie, we see a much lower MAE of 9.38\% on average.  While these error rates are more reasonable, we are concerned that this may be an artifact of both models having a low F1 score overall. 
We conclude that using GPT-4 does not work very well as a source of silver labels to assess model performance on unlabeled data for the XLM-R KPE model as compared to our proposed disagreement scores approach.

\begin{table}[htbp]
\centering
\scalebox{0.7}{
\begin{tabular}{llllll}
\hline
\textbf{Language} & \textbf{F1 (silver label)} & \textbf{F1 (golden label)} & \textbf{MAE}\\
\hline
XLM-R-EN & 0.392 & 0.705 & 0.313 \\
XLM-R-ES & 0.368 & 0.659 & 0.291 \\
XLM-R-FR & 0.661 & 0.765 & 0.104 \\
XLM-R-IT & 0.378 & 0.571 & 0.193 \\
Curie-EN & 0.410 & 0.480 & 0.070 \\
Curie-ES & 0.306 & 0.441 & 0.135 \\
Curie-FR & 0.590 & 0.674 & 0.084 \\
Curie-IT & 0.298 & 0.384 & 0.086 \\
\hline
\end{tabular}}
\caption{Silver label for XLM-R and Curie.}
\label{tab:silver-labelling}
\end{table}

\vspace{-0.6cm}
\section{Conclusion}

We conclude that disagreement scoring is a promising approach to predict model performance of LLMs. LLMs like GPT-4 that use few-shot prompting as a source for silver labels have high MAE and may not be useful in practice. In this paper, we explored the effects over three LLM models, XLM-R, GPT-3, and GPT-4 across 10 languages and 3 domains. Overall we recommend against measuring model performance on complex NLP tasks using LLMs as a few-shot Oracle, in our experiments we observe GPT-4 derived labeling results in F1 prediction with MAE of 15.7\% on average (Table \ref{tab:silver-labelling}), with some MAE as high as 31.3\%. Instead we recommend using disagreement scores and related techniques,  from our experiments we observe MAE across various languages and domains to be 1.91\% on average, with some as high as 9\%. 

\section{Limitations}
We observe that the performance of our proposed GPT-based approaches work better on European languages than Asian languages.  We believe this could be improved upon by using different base LLMs that have been trained on more non-EU data and studying in more detail the trade-off of using more or less regression points to predict an unknown F1. Our experiments are also limited to a single but complex NLP task, KPE. We also note that the theoretical error bound of this approach in terms domain shift is not guaranteed, as described in \cite{kirsch2022note}.
In future work we hope to expand our study of these methods on additional models and tasks to further increase confidence and understand where these methods may fail and potentially work towards methods with stronger theoretical bounds.

\section{Ethics Statement}

In this section, we hope to address any ethical considerations that may arise regarding the use of our internal and private dataset. The dataset was labeled by an internal labeling team that was competitively compensated for their time. The data was sampled across a large variety of brands within each industry in order to limit biases that may exist in specific domains. Lastly, the data was doubly anonymized to redact any brand sensitive or personal identifiable information (PII): first by an internally developed text anonymization algorithm, and then by human annotators.

\bibliography{anthology,custom}
\bibliographystyle{acl_natbib}

\cleardoublepage
\renewcommand{\thesubsection}{\Alph{subsection}}
\setcounter{table}{0}
\renewcommand{\thetable}{A\arabic{table}}
\setcounter{equation}{0}
\renewcommand{\theequation}{A\arabic{equation}}

\section*{Appendices}\label{sec:appendix}
\subsection{Analysis of the relationship of F1 and model error}
\label{sec:f1_error}

In Section~\ref{sec:adapt} we defined test error in equation~\ref{eq:testerror} and disagreement in equation~\ref{eq:disagreement}.  We can define accuracy in a similar way,
\begin{equation}\label{eq:acc}
     Test^{acc}_{\mathcal{D}}(h) = \mathbb{E}_{\mathcal{D}}[h(X) = Y] = 1 - Test^{err}_{\mathcal{D}}(h)
\end{equation}
where we test for equivalence instead of non-equivalence.  In this case we can see that minimizing $Test^{err}$ is equivalent to maximizing $Test^{acc}$.  By definition in Section~\ref{sec:adapt} we know that agreement and disagreement have a similar relationship, so that replacing model error with model accuracy and disagreement with agreement, we can transfer the same relationship established in Theorem \ref{thm: GeneralizationDisagreement} to \emph{model accuracy} and \emph{model agreement}.

Now, with respect to F1 score.  If we consider the discrete approximation of accuracy to be $\frac{TP + TN}{TP + TN + FP + FN}$, where $TP,TN,FP,FN$ are true positive/negatives and false positive/negatives respectively, and F1 is a harmonic mean between precision ($\frac{TP}{TP+FP}$) and recall ($\frac{TP}{TP+FN}$), which is $\frac{2TP}{2TP + FP + FN}$.  Then we can conclude that any increase/decrease in F1 (i.e. increase/decrease in $TP$ or decrease/increase in $FP,FN$) will result in a corresponding increase/decrease in accuracy, all else being equal.  Consequently, if our method predicts with low error a higher/lower F1 score, we can conclude that the corresponding model accuracy will also be higher/lower.

\subsection{Data statistics}
Table \ref{tab:survey-data-statistics} denotes the number of training, validation and testing data for each language of anonymized survey responses.  The corpus has data from 10 languages, English (EN), Spanish (ES), French (FR), Italian (IT), German (DE), Dutch (NL), Portuguese (PT), Japanese (JA), Chinese (ZH) and Korean (KO).
\begin{table}[htbp]
  \centering
  \scalebox{0.8}{\begin{tabular}{|c|c|c|c|}
    \hline
    \textbf{Language} & \textbf{Training} & \textbf{Validation} & \textbf{Testing} \\
    \hline
    EN & 28,000 & 2,000 & 2,206 \\
    ES & 16,000 & 1,679 & 1,000 \\
    FR & 7,000 & 1,000 & 1,501 \\
    IT & 5,000 & 1,000 & 1,591 \\ 
    DE & 1,500 & 500 & 912 \\
    PT & 1,500 & 500 & 1,000 \\
    NL & 1,500 & 500 & 1,000 \\
    KO & 2,465 & 500 & 1,000 \\
    JA  & 4,004 & 1,000 & 2,000 \\
    ZH & 2,986 & 1,000 & 2,000 \\
    \hline
  \end{tabular}} 
    \caption{Data statistics of anonymized survey responses.}
  \label{tab:survey-data-statistics}
\end{table}

\begin{table}[htbp]
\centering
\scalebox{0.7}{
\begin{tabular}{llllll}
\hline
\textbf{Language} & \textbf{Seed} & \textbf{Average score} & \textbf{F1 from model} & \textbf{Fitted F1} \\
\hline
JA & 1& 0.523 & 0.554 & 0.519 \\
& 11 & 0.537 & 0.560 & 0.530 \\
& 111 & 0.539 & 0.561 & 0.532 \\
& 1111 & 0.548 & 0.568 & 0.540 \\
\hline
FR & 1 & 0.843 & 0.765 & 0.778 \\
& 11 & 0.833 & 0.769 & 0.771 \\
& 111 & 0.856 & 0.765 & 0.788 \\
& 1111& 0.855 & 0.763 & 0.787  \\
\hline
KO & 1 & 0.776 & 0.717 & 0.724\\
& 11 & 0.764 & 0.716 & 0.715 \\
&111 & 0.773 & 0.706 & 0.722  \\
& 1111 & 0.771 & 0.717 & 0.721  \\
\hline
\end{tabular}}
\caption{XLM-R language change}
\label{tab:BERT-language-JA-FR-KO}
\end{table}

\subsection{Language Change for LLM}

In this section, we reported the detailed results for each testing language of XLM-R, Curie, and GPT-4 models in Tables \ref{tab:BERT-language-JA-FR-KO}, \ref{tab:Curie-language-JA-FR-KO}, and \ref{tab:GPT-4-language}. 
For each table, we show the agreement scores of different seeds in the third column, and corresponding F1 scores from the models in fourth column, and corresponding fitted F1 scores predicted from the linear function in fifth column. 

Note that for the results of Curie in Tables \ref{tab:Curie-language-JA-FR-KO}. In first three rows, we use the data points collected from EU and Asian languages to fit linear regression function and compute the performance. In row 4 and 5, we report the performance using prediction function based on data points from EU languages only.

\begin{table}[htbp]
\centering
\scalebox{0.7}{
\begin{tabular}{lllllll}
\hline
\textbf{Language}&\textbf{Seed} & \textbf{Average score} & \textbf{F1 from model} & \textbf{Fitted F1} \\
\\
\hline
JA (EU + Asian)&1 & 0.449 & 0.160 & 0.437 \\
&11 & 0.460 & 0.160 & 0.449\\
&111 & 0.463 & 0.160 & 0.451\\
&1111 & 0.469 & 0.160 & 0.457\\
\hline
FR (EU + Asian)& 1 & 0.618 & 0.675 & 0.580  \\
& 11 & 0.618 & 0.670 & 0.580 \\
& 111 & 0.618 & 0.676 & 0.580 \\
& 1111 & 0.608 & 0.677 & 0.570\\
\hline
KO (EU + Asian)&1 & 0.379 & 0.370 & 0.296 \\
&11 & 0.383 & 0.400 & 0.301 \\
&111 & 0.386 & 0.410 & 0.305 \\
&1111 & 0.397 & 0.400 & 0.319 \\
\hline
FR (EU only) & 1 & 0.618 & 0.675 & 0.641\\
&11 & 0.618 & 0.670 & 0.641\\
&111 & 0.618 & 0.675 & 0.641\\
&1111 & 0.608 & 0.677 & 0.632\\
\hline
ES (EU only)&1 & 0.413 & 0.443 & 0.446 \\
&11 & 0.410 & 0.442 & 0.443 \\
&111 & 0.410 & 0.444 & 0.443 \\
&1111 & 0.409 & 0.437 & 0.442 \\
\hline
\end{tabular}}
\caption{Curie language change}
\label{tab:Curie-language-JA-FR-KO}
\end{table}

\begin{table}[htbp]
\centering
\scalebox{0.7}{
\begin{tabular}{lllllll}
\hline
\textbf{Language}&\textbf{Seed} & \textbf{Average score} & \textbf{F1 from model} & \textbf{Fitted F1}\\
\hline
EN & 1 & 0.427   & 0.597  & 0.170 \\
&11 & 0.429   & 0.596  & 0.166 \\
&111 & 0.426   & 0.595  & 0.169 \\
&1111 & 0.425   & 0.592  & 0.166 \\
\hline
ES &1 & 0.325   & 0.305 &0.200 \\
&11 & 0.316   & 0.300 &0.016 \\
&111 & 0.320   & 0.302  & 0.017 \\
&1111 & 0.315   & 0.298  &0.017 \\
\hline
FR&1 &  0.604   & 0.428  & 0.031 \\
&11 & 0.595   & 0.426 & 0.028 \\
&111 & 0.592   & 0.426  & 0.027 \\
&1111 & 0.594  & 0.426 & 0.028 \\
\hline
IT&1 &  0.350   & 0.370 & 0.020 \\
&11 & 0.354   & 0.374  & 0.019 \\
&111 & 0.357   & 0.373  & 0.016 \\
&1111 & 0.364   & 0.374  & 0.009 \\
\hline
\end{tabular}}
\caption{GPT-4 Language change}
\label{tab:GPT-4-language}
\end{table}

\begin{table}[htbp]
\centering
\scalebox{0.7}{
\begin{tabular}{lllllll}
\hline
\textbf{Dataset} & \textbf{Seed} & \textbf{Average score} & \textbf{F1 from model} & \textbf{Fitted F1}\\
\hline
Conv &1 & 0.725 & 0.664 & 0.676 \\
&11 & 0.722 & 0.648 & 0.673\\
&111 & 0.735 & 0.683 & 0.685 \\
&1111 & 0.690 & 0.596 & 0.644\\
\hline
Twitter & 1 & 0.498 & 0.382 & 0.468 \\
&11 & 0.510 & 0.382 & 0.479 \\
&111 & 0.462 & 0.383 & 0.435 \\
&1111 & 0.4566 & 0.335 & 0.429 \\
\hline
\end{tabular}}
\caption{XLM-R domain change for conv and Twitter.}
\label{tab:BERT-modality-twitter}
\end{table}

\begin{table}[htbp]
\centering
\scalebox{0.7}{
\begin{tabular}{lllllll}
\hline
\textbf{Dataset} & \textbf{Seed} & \textbf{Average score} & \textbf{F1 from model} & \textbf{Fitted F1}\\
\hline
Conv &1 & 0.218 & 0.271 & 0.253 \\
&11 & 0.241 & 0.307 & 0.275\\
&111 & 0.209 & 0.274 & 0.244 \\
&1111 & 0.216 & 0.294 & 0.251\\
\hline
Twitter & 1 & 0.236 & 0.222 & 0.270 \\
&11 & 0.236 & 0.201 & 0.270 \\
&111 & 0.241 & 0.209 & 0.275 \\
&1111 & 0.237 & 0.210 & 0.271 \\
\hline
\end{tabular}}
\caption{Curie domain change for conv and Twitter.}
\label{tab:curie-modality-twitter-conv-Euro}
\end{table}

\begin{table}[htbp]
\centering
\scalebox{0.7}{
\begin{tabular}{lllllll}
\hline
\textbf{Dataset} & \textbf{Seed} & \textbf{Average score} & \textbf{F1 from model} & \textbf{Fitted F1}\\
\hline
Conv &1 & 0.552 & 0.368 & 0.474\\
&11 & 0.554 & 0.367 & 0.476\\
&111 & 0.555 & 0.364 & 0.477\\
&1111 & 0.560 & 0.375 & 0.480 \\
\hline
Twitter &1 & 0.531 & 0.299 & 0.459\\
&11 & 0.522 & 0.289 & 0.452\\
&111 & 0.536 & 0.293 & 0.463\\
&1111 & 0.539 & 0.288 & 0.465\\
\hline
\end{tabular}}
\caption{Curie domain change for conv and Twitter.}
\label{tab:GPT-4-modality-conv-twitter}
\end{table}

\subsection{Domain Change for LLM}
In this section, we reported the detailed results of domain change of XLM-R, GPT-3, and GPT-4 models in Tables \ref{tab:BERT-modality-twitter}, \ref{tab:curie-modality-twitter-conv-Euro}, and \ref{tab:GPT-4-modality-conv-twitter}. To be mentioned here, for Curie model performance in conv and Twitter domains, we use the data points collected from EU languages only due to the function shift with the introduction of Asian languages.

\begin{table}[htbp]
\centering
\scalebox{0.75}{
\begin{tabular}{llllll}
\hline
\textbf{Language} & \textbf{Seed} & \textbf{Predicted F1} & \textbf{F1 (golden)} \\
\hline
EN & 1&  0.390 & 0.710 \\
& 11 &  0.392 & 0.705 \\
& 111  & 0.397 & 0.710 \\
& 1111  & 0.389 & 0.697 \\
\hline
ES & 1  & 0.369 & 0.660 \\
& 11  & 0.368 & 0.658 \\
& 111  & 0.368 & 0.659 \\
& 1111 & 0.368 & 0.669  \\
\hline
FR & 1  & 0.657 & 0.765\\
& 11  & 0.666 & 0.769 \\
&111  & 0.659 & 0.765  \\
& 1111  & 0.661 & 0.763  \\
\hline
IT & 1  & 0.379 & 0.571\\
& 11 &  0.382 & 0.571 \\
&111 &   0.373 & 0.571  \\
& 1111  & 0.380 & 0.573  \\
\hline
\end{tabular}}
\caption{GPT-4 silver label for XLM-R}
\label{tab:GPT-4-silver-XLM-R}
\end{table}

\begin{table}[htbp]
\centering
\scalebox{0.75}{
\begin{tabular}{llllll}
\hline
\textbf{Language} & \textbf{Seed} & \textbf{Predicted F1} & \textbf{F1 (golden)} \\
\hline
EN & 1&  0.410 & 0.476 \\
& 11 &  0.410 & 0.485 \\
& 111  & 0.410 & 0.481 \\
& 1111  & 0.411 & 0.480 \\
\hline
ES & 1  & 0.305 & 0.443 \\
& 11  & 0.309 & 0.442 \\
& 111  & 0.308 & 0.443 \\
& 1111 & 0.304 & 0.437  \\
\hline
FR & 1  & 0.591 & 0.675\\
& 11  & 0.594 & 0.670 \\
&111  & 0.586 & 0.675  \\
& 1111  & 0.590 & 0.677  \\
\hline
IT & 1  & 0.297 & 0.386\\
& 11 &  0.298 & 0.382 \\
&111 &   0.299 & 0.387  \\
& 1111  & 0.297 & 0.382  \\
\hline
\end{tabular}}
\caption{GPT-4 silver label for XLM-R}
\label{tab:GPT-4-silver-Curie}
\end{table}

\subsection{GPT-4 prompt engineering silver label for
XLM-R and Curie}\label{sec:gpt-4-prompt}
In this section, we reported the detailed performance of GPT-4 models as a sliver label source for XLM-R and Curie models, and the results are shown in Table \ref{tab:GPT-4-silver-XLM-R} and \ref{tab:GPT-4-silver-Curie}. For the GPT-4 100-shot prompting, we randomly sample 100 data records from the anonymized survey data for each language, and then we follow the instructions \footnote{https://platform.openai.com/docs/guides/gpt/chat-completions-api} and use the Chat-completions functions. For the 100 random samples, we provide the text and its corresponding list of keyphrases. Then we ask GPT-4 to output keyphrases for new input text data.

For each table in the third column, we use the GPT-4 generated label as ground truth labels to test the model performance. For the fourth column, we use human annotated label as ground truth labels to test the model performance.

\end{document}